# SUPERVISED CONTRASTIVE LEARNING FOR FINE-GRAINED CHROMOSOME RECOGNITION

Ruijia Chang
*Hangzhou City University*
Hangzhou, China
ruijiachang@outlook.com

Suncheng Xiang
*Shanghai Jiao Tong University*
Shanghai, China
xiangsuncheng17@sjtu.edu.cn

Chengyu Zhou
*Liaoning Normal University*
Liaoning, China
32067896@qq.com

Kui Su
*Hangzhou City University*
Hangzhou, China
suk@hzcu.edu.cn

Dahong Qian
*Shanghai Jiao Tong Univerisity*
Shanghai, China
Dahong.qian@sjtu.edu.cn

Jun Wang
*Hangzhou City University*
Hangzhou, China
wjcy19870122@163.com

**ABSTRACT**

Chromosome recognition is an essential task in karyotyping, which plays a vital role in birth defect diagnosis and biomedical research. However, existing classification methods face significant challenges due to the inter-class similarity and intra-class variation of chromosomes. To address this issue, we propose a supervised contrastive learning strategy that is tailored to train model-agnostic deep networks for reliable chromosome classification. This method enables extracting fine-grained chromosomal embeddings in latent space. These embeddings effectively expand inter-class boundaries and reduce intra-class variations, enhancing their distinctiveness in predicting chromosome types. On top of two large-scale chromosome datasets, we comprehensively validate the power of our contrastive learning strategy in boosting cutting-edge deep networks such as Transformers and ResNets. Extensive results demonstrate that it can significantly improve models' generalization performance, with an accuracy improvement up to +4.5%. Codes and pretrained models will be released upon acceptance of this work.

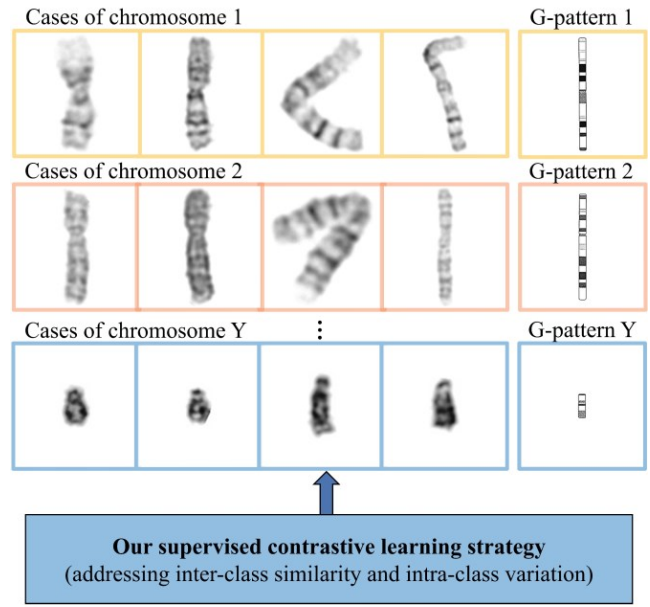

**Fig. 1.** Chromosome recognition is a highly challenging task due to substantial variations in sizes, shapes, G-band patterns, and image contrast, exhibiting significant inter-class similarity and intra-class variation. Our proposed supervised contrastive learning strategy can effectively address these challenges.

## 1. INTRODUCTION

The karyotype of a normal person comprises 22 pairs of autosomes and one pair of allosomes (XX or XY). Besides, every type of chromosome has a certain structure that can be identified through the Giemsa-stained band patterns (G-patterns) as depicted in Fig. 1. These G-patterns provide medical experts with valuable information for diagnosing genetic diseases during the karyotyping process[1]. Accurate chromosome classification is crucial for karyotyping, but it poses a significant challenge due to the complicated characteristics of chromosomes. As shown in Fig. 1, chromosome images exhibit substantial variations in sizes, shapes, G-patterns, and image contrast, leading to significant inter-class similarity and intra-class variation.

In recent year, various studies have explored deep learning approaches for chromosome classification[2]–[6] and instance segmentation[16]. Although the studies have made significant contributions to the chromosome recognition task, their methods may still face great challenges in fine-grained chromosome recognition, particularly due to the inter-class similarity and intra-class variation. Models trained with regular optimization strategies may fail to capture common patterns and intrinsic relationships within chromosomes, resulting in poor generalization performance across datasets from different hospitals or devices.

In this paper, to address the challenges posed by the inter-class similarity and intra-class variation, we propose a novel Supervised Contrastive Learning (SCL) strategy to train chromosome classifiers. Similar to existing contrastive learning methods[7], our SCL strategy involves designing a novel loss function that encourages matched samples to have closer embeddings while pushes apart those of mismatched samples. Through this strategy, fine-grained embeddings can be extracted to expand the inter-class boundaries and decrease the intra-class variations, making them more discriminative in predicting chromosome types. More importantly, the proposed training strategy is model-agnostic. It can be applied to any cutting-edge deep classifiers such as the Convolutional Neural Networks (CNNs)[8,17-18] and Vision Transformers[9]. This flexibility enables exploring the strengths of different networks thereby facilitating the choice of a more powerful and suitable approach for chromosome classification.

To validate the effectiveness of our SCL method, extensive experiments were conducted on two large-scale datasets. The results demonstrated that our SCL method can significantly improve model's generalization ability.

## 2. RELATED WORK

### 2.1. Contrastive Learning

Contrastive learning was initially developed to tackle the challenge of self-supervised learning, where the goal is to derive meaningful representations from unlabeled data[10]. This method involves creating pairs of similar and dissimilar samples and optimizing the feature space to bring similar samples closer while pushing dissimilar samples apart. In computer vision, contrastive learning has proven successful in tasks like image classification, object detection, and semantic segmentation by generating feature embeddings that capture intrinsic relationships within samples, improving downstream task accuracy. Recently, contrastive learning has also shown promise in training large vision models using natural language supervision. For example, Radford et al.[11] introduced CLIP for contrastive language-image pre-training, achieving robust zero-shot image classification. In our study, we demonstrate that contrastive learning is a valuable tool for learning reliable chromosomal representations.

### 2.2. Chromosome Recognition

Many studies have made significant strides in the domain of chromosome recognition tasks. For instance, Gupta et al. [4] utilized a Siamese Network architecture to achieve accurate chromosome classification, with an average accuracy of 85.6%. Sharma *et al.* [5] introduced a preprocessing step before classification. In their approach, bent chromosomes were initially straightened through cropping and stitching followed by length normalization. This preprocessing step improved the classification performance, resulting in an accuracy score of 86.7%. Zhang *et al.* [12] presented an interleaved and multi-task network for simultaneous chromosome straightening and classification, achieving an average accuracy of 98.1%. Qin *et al.* [6] proposed the Varifocal-Net for classification of both chromosome type and polarity, with a classification accuracy of 98.9%. Although their methods have shown promising results in their studies, their models may face significant performance drop when applied to datasets from different hospitals. Our proposed contrastive learning strategy effectively tackles this issue.

## 3. METHOD

The primary objective of our study is to train deep models with robust classification capabilities for the fine-grained chromosome recognition, i.e., distinguishing the 1-22 autosomes and the XY allosomes. However, achieving this objective presents a formidable challenge due to the inherent inter-class similarity and intra-class variation among chromosomes. In response to this challenge, we propose the SCL strategy that encourages the models to learn reliable chromosome representations as illustrated in Fig. 2. In the following subsections, we first provide an overview of the task formulation, and then we delve into the details of the contrastive learning strategy and the loss functions.

### 3.1. Task Formulation and Notations

As illustrated in Fig. 2, during the training stage, two pairwise batches of chromosome images are constructed, each consisting of 24 samples corresponding to the 1-22 autosomes and the allosomes XY. These samples first undergo a series of data augmentations, including random rotations, flips, and contrast adjustments for perturbations. Subsequently, they are fed into the model encoder to obtain two sets of latent embeddings. These embeddings are adjusted by the contrastive learning module to capture shared features within types and distinctive features across types. Finally, the adjusted embeddings are decoded to produce an output layer with 24 softmax scores used to determine chromosome types. These scores are regressed to their respective ground-truth (GT) category labels. It is important to note that, for clarity, two models are depicted in Fig. 2, referred to as the Master and Slave. In practice, only one model is optimized during the training phase.

In this paper, we use the following notations for the convenience of description: $B_1 \in R^{N \times C \times H \times W}$ and $B_2 \in R^{N \times C \times H \times W}$ represent the two image batches, where $N = 24$, $C = 1$, $H = 224$, and $W = 224$ indicate the number of samples in each batch, the channel, the height, and the width of a chromosome image, respectively. The corresponding embeddings are denoted as $E_1 \in R^{N \times D}$ and $E_2 \in R^{N \times D}$, where $D$ is the dimensionality of each embedding, with a default value of 1024 in this study. The predicted scores are given by $P_1 \in R^{N \times 24}$ and $P_2 \in R^{N \times 24}$. In addition, $y_{type} = [1, 2, \ldots, X, Y]$ and $y_{gt} = [0, 1, \ldots, 22, 23]$ indicate chromosome types and their corresponding GT labels for each batch.

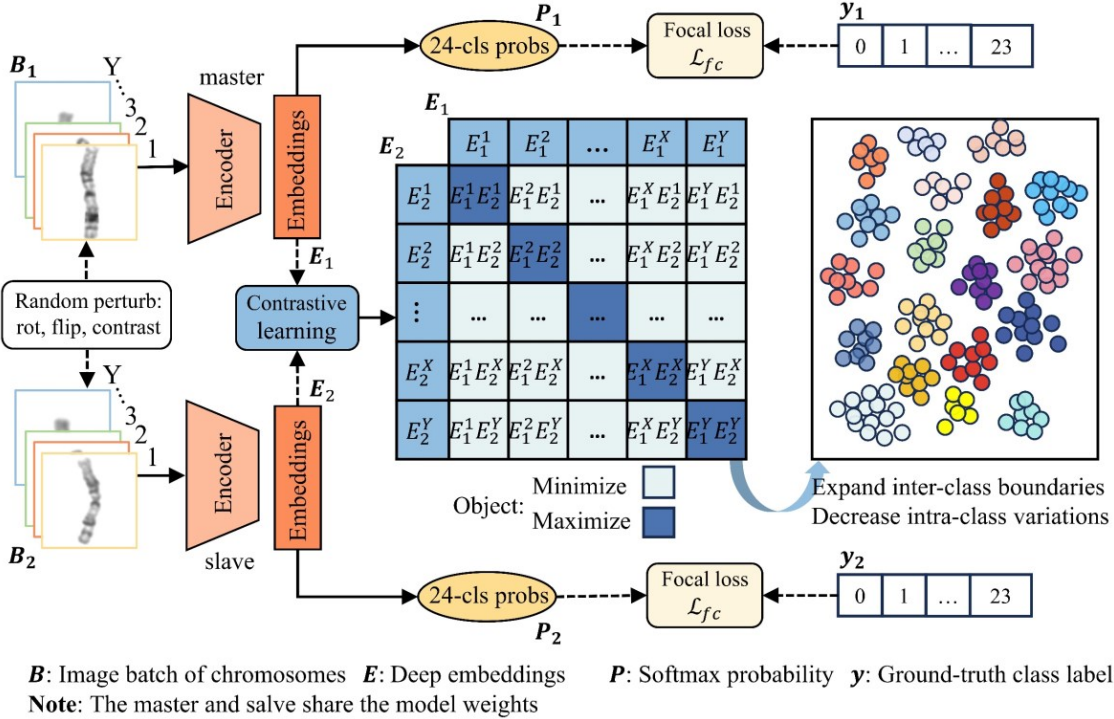

**Fig. 2.** Overall framework of our proposed supervised contrastive learning strategy. Two pairwise image batches are perturbated using random rotation, flipping, and contrast, and then fed into the model encoder to produce their deep latent embeddings. The contrastive learning is applied to force the embeddings more discriminative for class prediction. For clarity, two models, i.e., the Master and Slave, are depicted in the above framework. In fact, only one model is optimized during the training stage.

### 3.2. Supervised Contrastive Learning (SCL)

Let the model encoder be $f_e(\boldsymbol{\theta}_e)$ with the learnable parameters $\boldsymbol{\theta}_e$, the embeddings can be formulated as follows:

$$\boldsymbol{E}_i = f_e(\boldsymbol{B}_i|\boldsymbol{\theta}_e) \quad (1)$$

where $i \in \{1,2\}$ is the index of the pairwise image batches. To aid the model in learning shared features within types and distinctive features across types, we optimize the encoder to a set of parameters $\boldsymbol{\theta}_e^*$ that can maximize cosine similarity between the embeddings of matched pairs and minimize that of mismatched pairs as depicted in Fig. 2. Formally, this process can be expressed as follows:

$$\boldsymbol{\theta}_e^* = \begin{cases} \underset{\boldsymbol{\theta}}{\arg\max} \left( \sum_{i,j \in \{1,2\}} \sum_{k,l \in \boldsymbol{y}_{type}} E_{i\_norm}^k E_{j\_norm}^l \right) & \text{if } k = l \\ \underset{\boldsymbol{\theta}}{\arg\min} \left( \sum_{i,j \in \{1,2\}} \sum_{k,l \in \boldsymbol{y}_{type}} E_{i\_norm}^k E_{j\_norm}^l \right) & \text{otherwise} \end{cases} \quad (2)$$

where $E$ indicates an embedding in $\boldsymbol{E}_1$ or $\boldsymbol{E}_2$, and $E_{norm}$ is the $l2$-normalized embedding. $k$ and $l$ are the chromosome type of each sample. $E_{i\_norm}^k E_{j\_norm}^l$ denotes the inner product, i.e., the cosine similarity score, between the embedding $E_i^k$ and the embedding $E_j^l$. In practice, the above objective can be achieved by minimizing the following loss function:

$$\mathcal{L}_{con} = \sum_{i,j \in \{1,2\}} [\mathcal{L}_{fl}^{row}(\boldsymbol{S}_{ij}, \boldsymbol{y}_{gt}) + \mathcal{L}_{fl}^{col}(\boldsymbol{S}_{ij}, \boldsymbol{y}_{gt})], \quad (3)$$

where $\boldsymbol{S}_{ij} = \boldsymbol{E}_{i\_norm} \boldsymbol{E}_{j\_norm}$ means the matrix of cosine similarity between the normalized embeddings $\boldsymbol{E}_i$ and the embeddings $\boldsymbol{E}_j$ (see Fig. 2). $\mathcal{L}_{fl}^*$ denotes the focal loss[13] for multiple classification along the rows and the columns of the matrix. The NumPy-like pseudo code of the SCL is summarized in Algorithm 1.

### 3.3. Total Loss Function

During the training phase, the following loss function is applied to optimize the whole classification process:

$$\mathcal{L}_{total} = \lambda_1 \mathcal{L}_{con} + \lambda_2 \mathcal{L}_{cls}, \quad (4)$$

where $\mathcal{L}_{cls}$ is the loss function for the classification output. It is calculated using the focal loss as follows:

**Algorithm 1** Pseudo code of SCL for an iteration

**Inputs**:
   Image batch 1: $B_1$ # shape (24, 1, 224, 224)
   Image batch 2: $B_2$ # shape (24, 1, 224, 224)
   GT labels: $y_{gt} = [0, 1, 2, ..., 22, 23]$ #shape (24)

\# Obtain embeddings from the inputs using the encoder
1: $E_1 = f_e(B_1 | \theta_e)$ # shape (24, 1024)
2: $E_2 = f_e(B_2 | \theta_e)$ # shape (24, 1024)

\# Perform $l_2$ normalization on the deep embeddings
3: $E_{1\_norm} = l_2\_norm(E_1)$ # shape (24, 1024)
4: $E_{2\_norm} = l_2\_norm(E_2)$ # shape (24, 1024)

\# Calculate cosine similarity matrices from the embeddings
\# $t$: a learnable temperature factor
5: $S_{11} = \text{dot}(E_{1\_norm}, E_{1\_norm}.\text{trans}()) \times t$ # shape (24, 24)
6: $S_{12} = \text{dot}(E_{1\_norm}, E_{2\_norm}.\text{trans}()) \times t$ # shape (24, 24)
7: $S_{22} = \text{dot}(E_{2\_norm}, E_{2\_norm}.\text{trans}()) \times t$ # shape (24, 24)

\# Calculate the focal losses based on the similarity matrices
8: $\mathcal{L}_{con} = \sum_{i,j \in \{1,2\}} [\mathcal{L}_{fl}^{row}(S_{ij}, y_{gt}) + \mathcal{L}_{fl}^{col}(S_{ij}, y_{gt})]$

\# Update parameters of the encoder using gradient descent
\# $\gamma$: learning rate
9: $\theta_e^* = \theta_e + \gamma \frac{\nabla \mathcal{L}_{con}}{\nabla \theta_e}$

## 4. EXPERIMENTS AND RESULTS

### 4.1. Datasets

Two large-scale datasets were adopted to develop and validate the proposed method: 1) A private dataset named *OGH-Fudan* with 120,378 G-band chromosome images of 553 individuals obtained from a cooperative hospital. Each image has a resolution of 256 × 256 pixels. The dataset was divided into three subsets: 106,260 images for training, 2,300 for validation, and 11,818 for testing; 2) a publicly available dataset named *ChrmNet*[14] for external validation. This dataset contains 126,453 privacy preserving G-band chromosome instances from 408 individuals. The image size is 300 × 300 pixels.

### 4.2. Training Details and Evaluation Metrics

To validate the effectiveness of our SCL strategy, we conducted ablation study by training several state-of-the-art (SOTA) image classifiers with or without using the SCL strategy. These classifiers were implemented in the TIMM[1] library, which is a comprehensive open-source Python library designed for various computer vision tasks, offering a wide range of pre-trained models. All models were trained using Adam optimizer[15] with an initial learning rate of 1e-5 for a total of 200 epochs, with each epoch consisting of 2000 steps. Moreover, we compared to two SOTA chromosome classifiers: the Varifocal-Net[6] and the HR-Net[12]. The classification performance was evaluated using the following

$$\mathcal{L}_{cls} = \mathcal{L}_{fl}(P_1, y_{gt}) + \mathcal{L}_{fl}(P_2, y_{gt}). \quad (5)$$

The $\lambda_1$ and $\lambda_2$ in Eq. (4) are two hyperparameters used to adjust the contribution of each loss term. In this study, both $\lambda_1$ and $\lambda_2$ are empirically set to a default value of 1.0.

| | | *OGH-Fudan* | | | *ChrmNet* | | |
|---|---|---|---|---|---|---|---|
| Classifiers | With SCL | *Accuracy* (%) | *Recall* (%) | *AUC* (%) | *Accuracy* (%) | *Recall* (%) | *AUC* (%) |
| Vit_small | ☒ | 91.64 | 91.27 | 99.73 | 76.62 | 76.66 | 97.11 |
| Vit_small | ☑ | 96.17(+4.53) | 95.54(+4.27) | 99.88(+0.15) | 84.28(+7.66) | 84.01(+7.35) | 98.82(+1.71) |
| Vit_large | ☒ | 91.57 | 91.07 | 99.69 | 75.26 | 75.00 | 96.35 |
| Vit_large | ☑ | 93.34(+1.77) | 92.65(+1.58) | 99.70(+0.01) | 78.37(+3.11) | 78.03(+3.03) | 97.38(+1.03) |
| Davit_small | ☒ | 90.20 | 90.06 | 99.62 | 73.68 | 73.79 | 97.32 |
| Davit_small | ☑ | 94.13(+3.93) | 93.53(+3.47) | 99.77(+0.15) | 81.18(+7.50) | 80.86(+7.07) | 98.33(+1.01) |
| Davit_large | ☒ | 92.58 | 92.18 | 99.74 | 77.38 | 77.18 | 97.73 |
| Davit_large | ☑ | 94.54(+1.96) | 93.71(+1.53) | 99.80(+0.06) | 82.01(+4.63) | 81.84(+4.66) | 98.34(+0.61) |
| Swin_small | ☒ | 93.65 | 93.18 | 99.80 | 81.60 | 81.48 | 98.74 |
| Swin_small | ☑ | 96.07(+2.42) | 95.46(+2.28) | 99.88(+0.08) | 86.51(+4.91) | 86.16(+4.68) | 99.08(+0.34) |
| Swin_large | ☒ | 95.30 | 94.87 | 99.86 | 83.22 | 83.05 | 98.87 |
| Swin_large | ☑ | **96.33(+1.03)** | **95.71(+0.84)** | **99.88(+0.02)** | **87.32(+4.10)** | **86.98(+3.93)** | **99.24(+0.37)** |
| ResNet-34 | ☒ | 93.46 | 93.25 | 99.80 | 82.49 | 82.62 | 98.66 |
| ResNet-34 | ☑ | 95.22(+1.76) | 94.94(+1.69) | 99.86(+0.06) | 86.38(+3.89) | 86.41(+3.79) | 99.14(+0.48) |
| ResNet-50 | ☒ | 89.64 | 89.49 | 99.61 | 76.58 | 76.95 | 98.13 |
| ResNet-50 | ☑ | 91.95(+2.31) | 91.50(+2.01) | 99.73(+0.12) | 80.32(+3.74) | 80.44(+3.49) | 98.42(+0.29) |

**Table 1.** Ablation study. Results of classifiers for the chromosome recognition task. The best performance is shown in bold. SCL: the proposed supervised contrastive learning. *OGH-Fudan*: internal validation. *ChrmNet*: external validation. Vit_small: Vit_small_patch8_224. Vit_large: Vit_large_patch14_224. Swin_small: Swin_small_patch4_window7_224. Swin_large: Swin_large_patch4_window7_224.

---

[1] https://github.com/huggingface/pytorch-image-models

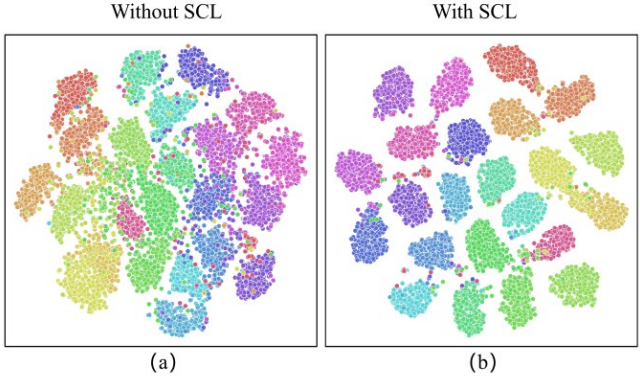

**Fig. 3.** The t-SNE plots represent the learned features of chromosomes using the Vit Transformer (Vit_small_patch8_224), with (a) the absence of SCL strategy, and (b) the incorporation of SCL strategy.

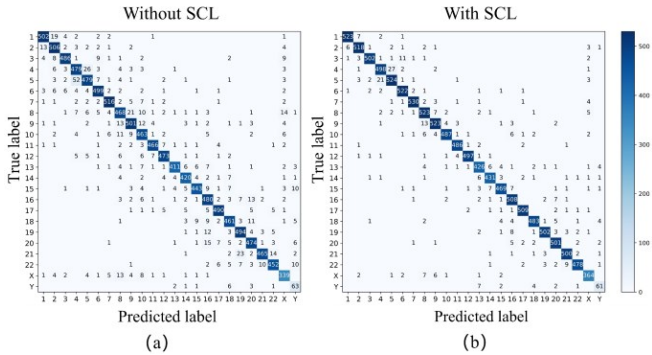

**Fig. 4.** Confusion matrix of the Vit_small_patch8_224 model applied to chromosomes, distinctly illustrating the impact of the SCL strategy: (a) in the absence of the SCL strategy, and (b) with the incorporation of the SCL strategy.

metrics: *Accuracy*, *Recall*, and Area under the ROC curve (*AUC*).

### 4.3. Results and Analysis

Table 1 presents the results of ablation study, which demonstrate the effectiveness of our SCL strategy in improving the models' performance, especially for smaller models. The *Accuracy* scores of all small models are improved by more than 2.4% on the *OGH-Fudan* dataset during the internal validation. Interestingly, the gaps are increased significantly when using the *ChrmNet* dataset for the external validation. For example, the Vit_small_patch8_224 model achieves improvements of 7.66%, 7.35%, and 1.71% in *Accuracy*, *Recall*, and *AUC* scores, respectively, when using the SCL strategy. In contrast, the counterparts achieved during the internal validation are only 4.53%, 4.27%, and 0.15%, respectively. These results

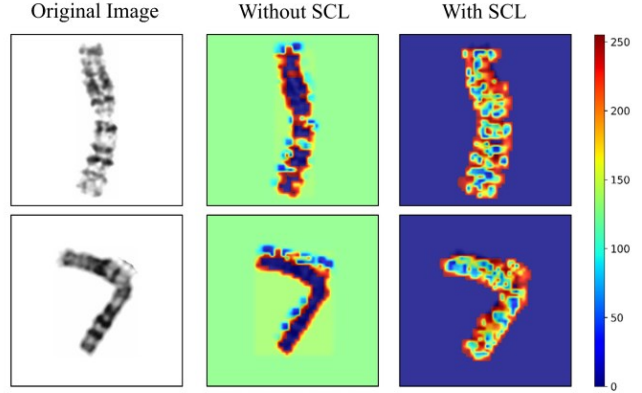

**Fig. 5.** The original images and heatmaps of two chromosomes processed by the same layer of the ResNet50 model. The heatmaps represent the chromosomal features, with one heatmap showing the features processed without the SCL strategy and the other with the SCL strategy.

indicate that our SCL strategy can effectively enhance models' generalization ability.

To further illustrate the impact of our SCL strategy, we present a t-SNE visualization comparing the differences between the embeddings of the Vit_small_patch8_224 model with SCL and without SCL (see Fig. 3). Obviously, the data points are more distinctly separated in the t-SNE plot when SCL is employed, indicating better discriminative capabilities of the embeddings. In contrast, the plot without SCL shows more overlapping and ambiguous boundaries between different classes. This visual representation highlights the ability of SCL to enhance the model's ability to learn meaningful representations and improve classification performance.

In addition, we plot a confusion matrix to analyze which classes are more impacted by the SCL strategy (see Fig. 4). The matrix shows that the SCL strategy can significantly reduce misclassifications, particularly for those classes exhibiting severe inter-class similarity. For instance, the Vit_small_patch8_224 model without the SCL has 52 misclassifications between the 4th and 5th autosomes. These misclassifications are reduced to 21 after applying the SCL strategy.

Moreover, we provide heatmaps to demonstrate the significant enhancement in the model's focus on chromosomal structures when the SCL strategy is applied (see Fig. 5). In these heatmaps, the color intensity reflects the model's attention to different features of the chromosomes, where blue indicates low attention and red high attention. As shown in the heatmaps, without the SCL strategy, the model tends to concentrate more on the edges of the chromosomes, especially when processing curved chromosome images. However, with the SCL strategy, there is a notable shift in the model's focus. It begins to pay more attention to the internal structures and striations of the chromosomes, which often contain critical information for identifying the type of chromosomes.

|  | OGH-Fudan | | | ChrmNet | | |
| --- | --- | --- | --- | --- | --- | --- |
| Classifiers | Accuracy (%) | Recall (%) | AUC (%) | Accuracy (%) | Recall (%) | AUC (%) |
| Swin_large_patch4_window7_224 | **96.33** | **95.71** | **99.88** | **87.32** | **86.98** | **99.24** |
| Varifocal-Net[6] | 93.18 | 92.81 | 99.82 | 80.83 | 81.06 | 98.50 |
| HR-Net[12] | 91.34 | 91.21 | 99.69 | 77.99 | 78.07 | 98.13 |

**Table 2.** Comparison with two SOTA chromosome classifiers. The best performance is shown in bold. *OGH-Fudan*: internal validation. *ChrmNet*: external validation.

Finally, we tabulate the results of our Swin_large model (the best one in Table 1) and compare them with the SOTA chromosome classifiers (i.e., the Varifocal-Net and the HR-Net) in Table 2. Our method outperforms these methods in terms of all evaluation metrics. A similar phenomenon to Table 1 can be observed that the performance of all methods decreases significantly on external validation. However, our method remains the best one with much higher *Accuracy* and *Recall* scores.

## 5. CONCLUSION

In this paper, we focus on tackling difficulties in chromosome classification, particularly the issues of inter-class similarity and intra-class variation. We introduced a model-agnostic approach based on contrastive learning to effectively boost models' performance and generalization ability in out-of-distribution datasets. Through a comprehensive series of experiments, which included ablation studies and comparisons with other chromosome classifiers, we have demonstrated the effectiveness of our method.